\documentclass{article}

\usepackage[preprint]{neurips_2025}
\usepackage{hyperref}

\usepackage[table]{xcolor}

\definecolor{customblue}{rgb}{0.2, 0.3, 0.8}
\definecolor{customgreen}{rgb}{0.1, 0.6, 0.3}

\definecolor{lightblue}{RGB}{173, 216, 230}
\definecolor{lightgray}{gray}{0.9}
\definecolor{lightgreen}{RGB}{144, 238, 144}
\definecolor{lightred}{RGB}{255, 182, 193}

\definecolor{wkred}{RGB}{255, 190, 190}
\definecolor{wkblue}{RGB}{210, 230, 250}
\definecolor{wkgreen}{RGB}{226,240,217}

\definecolor{skyblue}{RGB}{0, 102, 204}

\usepackage{listings}
\usepackage{multirow}
\usepackage{tabularx}
\usepackage{booktabs}

\usepackage{float}

\usepackage{etoolbox, subcaption}
\usepackage[font=footnotesize,labelfont=bf,aboveskip=1pt,belowskip=-10pt]{caption}

\newdimen\abovecrulesep
\newdimen\belowcrulesep
\makeatletter
\patchcmd{\@@@cmidrule}{\aboverulesep}{\abovecrulesep}{}{}
\patchcmd{\@xcmidrule}{\belowrulesep}{\belowcrulesep}{}{}

\definecolor{demphcolor}{RGB}{144, 144, 144}
\definecolor{mygray}{gray}{0.4}
\definecolor{lightgray}{rgb}{0.9, 0.9, 0.9}

\newlength\savewidth

\newcommand{\tablestyle}[2]{\setlength{\tabcolsep}{#1}\renewcommand{\arraystretch}{#2}\centering\footnotesize}
\makeatletter\renewcommand\paragraph{\@startsection{paragraph}{4}{\z@}{.5em\@plus1ex\@minus.2ex}{-.5em}{\normalfont\normalsize\bfseries}}
\makeatother

\newcolumntype{C}[1]{>{\centering\arraybackslash}p{#1}}
\newcolumntype{R}[1]{>{\raggedleft\arraybackslash}p{#1}}
\newcolumntype{L}[1]{>{\raggedright\arraybackslash}p{#1}}

\usepackage[utf8]{inputenc} 
\usepackage[T1]{fontenc}    

\usepackage{url}            
\usepackage{booktabs}       
\usepackage{amsfonts}       
\usepackage{nicefrac}       
\usepackage{microtype}      
\usepackage{xcolor}         

\usepackage{graphicx}
\graphicspath{{assets/}} 
\usepackage{amsmath}
\usepackage{amssymb}
\usepackage{booktabs}
\usepackage{enumitem}
\usepackage{algorithm}
\usepackage{algpseudocode}
\usepackage{bm}
\usepackage{multirow}
\usepackage{caption}
\usepackage{subcaption}
\usepackage{tikz}
\usepackage{tabularx}

\usepackage{arydshln}

\usepackage{adjustbox}
\usepackage{natbib}
\definecolor{darkGreen}{RGB}{0,100,0} 

\usepackage{pifont}

\usepackage{lipsum}
\usepackage{bbm}
\usepackage{dsfont}

\makeatletter
\providecommand{\@notice}{}
\makeatother

\newcommand{\work}{\text{Temporal-RLT}}

\title{Reinforcement Learning Tuning for VideoLLMs: Reward Design and Data Efficiency}

\author{
    Hongyu Li\textsuperscript{1}\thanks{Equal contribution.} , 
    Songhao Han\textsuperscript{1}\footnotemark[1] ,
    Yue Liao\textsuperscript{2}\footnotemark[1]  \thanks{Project leader.} ,
    Junfeng Luo\textsuperscript{3},
    \textbf{Jialin Gao\textsuperscript{3},} \\
    \textbf{Shuicheng Yan\textsuperscript{2},}
    \textbf{Si Liu\textsuperscript{1}\thanks{Corresponding author.}} \\
    \\$^{1}$ BUAA \quad $^{2}$ NUS \quad $^{3}$ Meituan  \\
}

\begin{document}

\maketitle

\begin{figure}[ht]
   \centering
   \vspace{-3em}
   \includegraphics[width=1.0\textwidth]{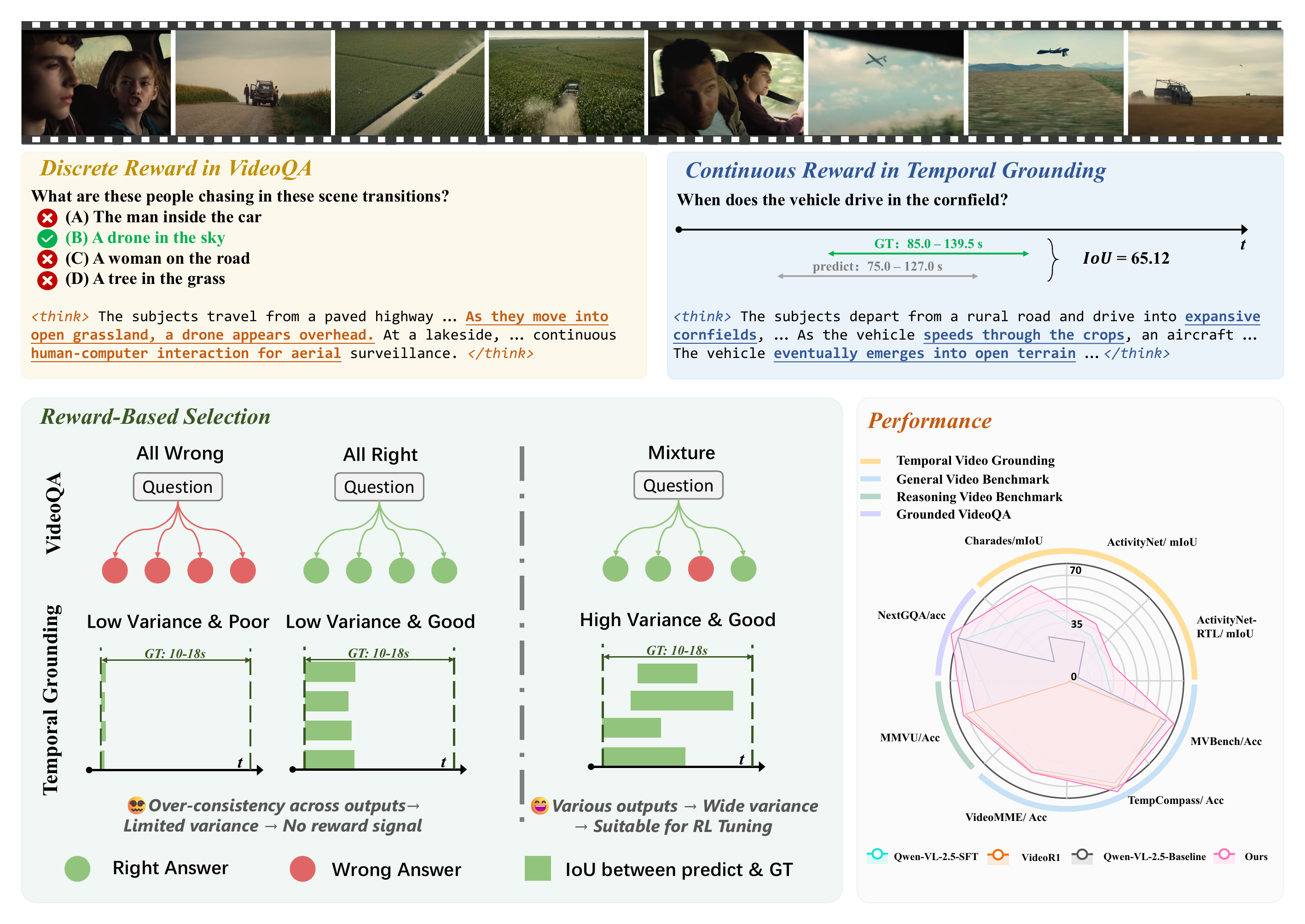} 
   \caption{\textbf{Overview of our reward-based framework for post-training on video understanding tasks.} 
\textit{Top left:} Semantic reasoning is supervised using a discrete reward from multi-choice VideoQA, where the model selects the correct answer. 
\textit{Top right:} Temporal reasoning is guided by a continuous reward based on temporal IoU in grounding tasks. 
\textit{Bottom left:} A reward-based data selection strategy filters training samples based on behavioral variance. High-variance samples induce stronger preference gradients and are prioritized for GRPO optimization. 
\textit{Bottom right:} Performance comparison across baselines and our Temporal-RLT, showing consistent improvements on QA, grounding, and reasoning tasks.}
   \label{fig:fig1}
\end{figure}

\begin{abstract}
Understanding real-world videos with complex semantics and long temporal dependencies remains a fundamental challenge in computer vision. Recent progress in multimodal large language models (MLLMs) has demonstrated strong capabilities in vision-language tasks, while reinforcement learning tuning (RLT) has further improved their reasoning abilities. In this work, we explore RLT as a post-training strategy to enhance the video-specific reasoning capabilities of MLLMs. Built upon the Group Relative Policy Optimization (GRPO) framework, we propose a dual-reward formulation that supervises both semantic and temporal reasoning through discrete and continuous reward signals. To facilitate effective preference-based optimization, we introduce a variance-aware data selection strategy based on repeated inference to identify samples that provide informative learning signals. We evaluate our approach across eight representative video understanding tasks, including VideoQA, Temporal Video Grounding, and Grounded VideoQA. Our method consistently outperforms supervised fine-tuning and existing RLT baselines, achieving superior performance with significantly less training data. These results underscore the importance of reward design and data selection in advancing reasoning-centric video understanding with MLLMs. Notably, The initial code release (two months ago) has now been expanded with updates, including optimized reward mechanisms and additional datasets. The latest version is available at \href{https://github.com/appletea233/Temporal-R1}{\textit{\textcolor{cyan!100!gray!50!}{Temporal-RLT}}}.
\end{abstract}

\section{Introduction}
Video is a fundamental medium for capturing the dynamics of the visual world. Understanding real-world videos—characterized by complex semantics, diverse visual content, and long-range temporal dependencies—remains a core challenge in computer vision. While large multimodal models (MLLMs) have shown promising capabilities in vision-language tasks, recent efforts have begun extending them to video understanding~\cite{cheng2024videollama,wang2025internvideo25,qwen25vl,li2024llava,zhang2024long,li2025llava}, where temporal alignment and structured reasoning are critical. Inspired by human cognitive processes, incorporating explicit reasoning into large language and multimodal models has been shown to significantly enhance their comprehension, particularly in tasks requiring multi-step inference and temporal understanding. Reinforcement learning tuning (RLT)~\cite{guo2025deepseek,shao2024deepseekmath,visualrft,videor1,shen2025vlm} has emerged as an effective post-training paradigm for stimulating such reasoning abilities, with notable success in both language and vision-language domains through preference-based optimization.

In this work, we propose \textbf{Temporal-RLT,} a RLT framework designed to improve the video-specific reasoning capabilities of MLLMs. Our method builds upon the Group Relative Policy Optimization (GRPO) algorithm~\cite{shao2024deepseekmath}, which optimizes model behavior by comparing multiple sampled outputs and applying fine-grained, verifiable reward signals. While GRPO has been successfully applied in textual reasoning tasks, its extension to video domains remains largely underexplored and introduces new challenges in modeling structured temporal reasoning under multimodal supervision.

The first challenge is \emph{"how to design effective reward functions that incentivize video-specific reasoning."} To address this, as shown in Fig.~\ref{fig:fig1}, we propose a dual-reward formulation that combines discrete rewards from multi-choice VideoQA with continuous rewards derived from temporal Intersection over Union (tIoU) in temporal grounding tasks. The discrete reward promotes semantic alignment by supervising answer correctness, while the continuous reward guides temporal localization accuracy. We further extend this formulation to Grounded VideoQA, where the model must both answer questions and localize supporting video segments, thereby enhancing interpretability and temporal grounding.

The second challenge involves identifying \emph{"effective training samples for GRPO-based RLT"}. Since GRPO relies on behavioral variance among sampled outputs to generate preference gradients, we propose a variance-aware data selection strategy based on repeated inference. For each sample, we perform multiple forward passes using a base VideoLLM and measure prediction variability. In multi-choice QA, we retain samples with moderate answer consistency to ensure a mix of correct and incorrect responses. In temporal grounding, we compute the gap between maximum and average IoU to identify samples with high intra-group variance. This approach filters out uninformative examples and focuses training on samples most likely to produce meaningful learning signals under GRPO. 

Based on this strategy, we construct two datasets to support Temporal-RLT. \textbf{Temporal-RLT-Full-490k} serves as a comprehensive dataset covering VideoQA, Temporal Grounding, and Grounded VideoQA. From this data pool, we derive \textbf{Temporal-RLT-32k}, a high-quality subset selected for its diversity and optimization potential, enabling efficient training with significantly less data.

We validate Temporal-RLT across 8  benchmarks spanning General VideoQA, Reasoning VideoQA, Temporal Grounding, and Grounded VideoQA. Our method consistently outperforms strong supervised fine-tuning (SFT) baselines, and achieves superior performance compared to Video-R1~\cite{videor1}. Additionally, we conducted extensive ablation studies to verify the effectiveness of our data selection strategy and the performance gains achieved by various rewards across different tasks. These results highlight the effectiveness of our reward design and data selection strategy for enhancing video-specific reasoning, while demonstrating strong performance with data efficiency in RLT-based video understanding.

\section{Temporal-RLT}
\label{sec:cot}
\begin{figure}[t]
   \centering
   \includegraphics[width=1.0\textwidth]{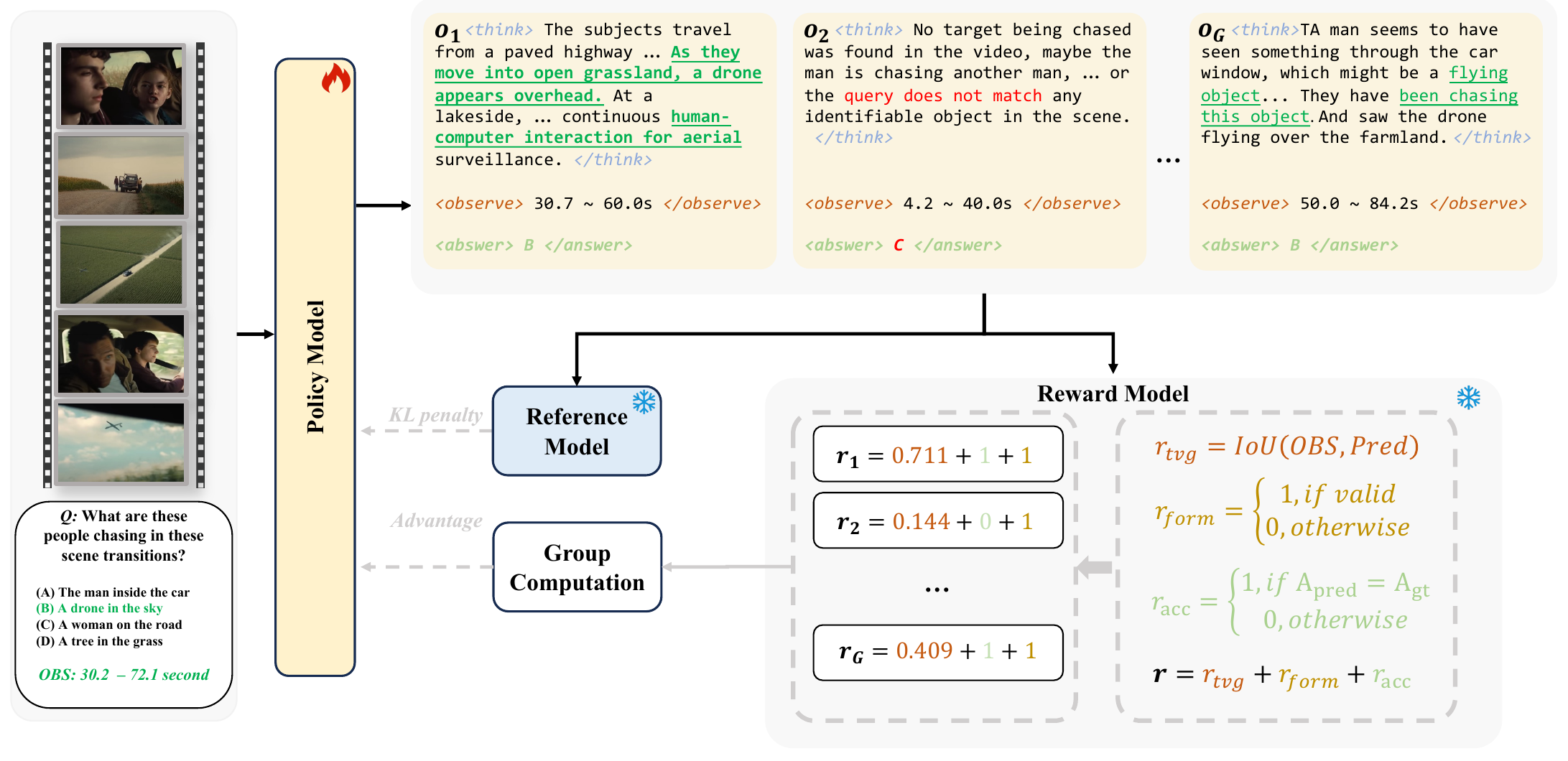} 
   \caption{\textbf{Overview of our training framework.} Given a video and its corresponding instruction, the VideoLLM generates either an answer choice for multi-choice QA or a temporal segment for grounding. The model is optimized using a GRPO-based RLT pipeline with task-specific rewards, including a discrete reward for QA and a continuous temporal IoU reward for grounding.}

   \label{fig:fig2}
\end{figure}

In this section, we present \emph{Temporal-RLT}, our reinforcement learning tuning pipeline tailored for video large language models (VideoLLMs). As illustrated in Fig.~\ref{fig:fig2}, Temporal-RLT extends a standard VideoLLM architecture by integrating a GRPO-based RLT module to further enhance the model's video-specific reasoning and comprehension capabilities. We begin by outlining the overall pipeline of the base VideoLLM (Sec.~\ref{sec:basemodel}). We then introduce our dual reward formulation (Sec.~\ref{sec:reward}), consisting of a discrete semantic reward from multi-choice QA (MC-QA) and a continuous temporal reward based on temporal IoU (tIoU). Finally, we describe how GRPO is incorporated into the Temporal-RLT framework under this dual-reward setting, forming a unified post-training strategy for video understanding (Sec.~\ref{sec:pre}).

\subsection{Base Model: QwenVL-2.5 with Structured Reasoning}
\label{sec:basemodel}

We adopt QwenVL-2.5~\cite{qwen25vl} as the base VideoLLM in our framework. QwenVL-2.5 is a powerful open-source multimodal language model pre-trained on large-scale video-language data, demonstrating strong performance across various video understanding tasks. Given multimodal input—comprising video frames and a natural language instruction—the model autoregressively generates outputs such as answer choices or grounding spans in a textual format. To move beyond direct answer prediction and promote explicit reasoning, we introduce a structured output format that encourages the model to \emph{think before answering}. Specifically, we define a standardized response template using \texttt{<think>} and \texttt{<answer>} tags:  
\[
\texttt{<think>}~\textit{reasoning trace}~\texttt{</think>}~\texttt{<answer>}~\textit{final response}~\texttt{</answer>}.
\]
This format enforces the model to articulate its reasoning process prior to providing the final answer, enabling more interpretable and consistent behavior during inference. To further incentivize structural adherence and output consistency, we introduce a format alignment reward \( R_{\text{format}} \), which encourages the model to generate outputs that conform to the predefined structure.

\subsection{Dual Verifiable Rewards in Video Tasks}
\label{sec:reward}

Reward design plays a critical role in guiding optimization within the GRPO-based RLT framework. To enhance both global semantic and fine-grain temporal reasoning in VideoLLMs, we introduce two complementary reward signals. A discrete semantic reward, derived from multi-choice QA tasks, encourages global semantic alignment. A continuous temporal reward, based on temporal IoU from grounding tasks, promotes fine-grained temporal comprehension. Together, these rewards provide structured supervision across multiple reasoning dimensions.

\textbf{Global Discrete Semantic Reward.}  
To incentivize high-level semantic reasoning, we leverage the Multi-Choice Video Question Answering (MC-VQA) task. In this setting, the model is given a video $V$, a question $Q$, and a set of candidate answers $\{A_1, A_2, ..., A_n\}$. The objective is to select the correct answer $A_c$, where $c$ denotes the index of the ground truth option. This transforms open-ended reasoning into a classification problem over a constrained semantic space.

To provide a verifiable reward signal, we adopt a binary accuracy-based reward function:
\begin{equation}
\small
R_{\text{acc}} = 
\begin{cases} 
1, & \text{if } A_{\text{pred}} = A_{\text{gt}} \\
0, & \text{otherwise}
\end{cases}
\end{equation}
where $A_{\text{pred}}$ is the model's selected answer and $A_{\text{gt}}$ is the ground truth. This discrete reward guides the model to align its global semantic understanding with task-specific supervision.

\textbf{Fine-Grained Temporal IoU Reward.}  
To capture fine-grained temporal reasoning, we incorporate the Temporal Video Grounding (TVG) task, where the model is required to localize an event within the video timeline. Given a video $V$ and a language query, the model predicts a temporal segment defined by start and end timestamps $(S_p, E_p)$, which is then compared against the ground truth interval $(S_g, E_g)$.

We define the reward as the temporal IoU between the predicted and ground truth segments:
\begin{equation}
\small
R_{\text{IoU}} = \frac{\max(0, \min(E_p, E_g) - \max(S_p, S_g))}{\max(E_p, E_g) - \min(S_p, S_g)}
\end{equation}
This continuous reward quantifies the alignment between predicted and reference segments, encouraging precise temporal localization. It provides a smooth gradient signal that complements the binary supervision from semantic QA, enabling the model to reason across multiple granularity levels.

\subsection{Temporal-RLT Optimization with GRPO}
\label{sec:pre}

To optimize VideoLLMs under our task-specific reward formulations, we adopt GRPO, a preference-based reinforcement learning algorithm that fine-tunes models without requiring an explicit value function or learned critic. Instead of estimating absolute returns, GRPO compares the quality of multiple responses sampled for the same input and uses their relative rankings to guide optimization. This makes GRPO naturally compatible with our verifiable and multi-dimensional reward design.

We define three reward functions tailored to distinct video reasoning tasks. For Multi-Choice VideoQA (MC-QA), the reward is defined as $R_{\text{mc}} = R_{\text{format}} + R_{\text{acc}}$, where $R_{\text{acc}}$ indicates answer correctness and $R_{\text{format}}$ enforces output consistency. For Temporal Video Grounding (TVG), we use $R_{\text{tvg}} = R_{\text{format}} + R_{\text{IoU}}$, where $R_{\text{IoU}}$ measures the alignment between predicted and ground-truth temporal segments. To bridge semantic and temporal reasoning, we further introduce Grounded VideoQA (GQA), where models are required to both answer the question and highlight relevant temporal spans using \texttt{<observe>} tags. The reward is defined as $R_{\text{gqa}} = R_{\text{format}} + \frac{1}{2}(R_{\text{acc}} + R_{\text{IoU}})$, encouraging joint semantic accuracy and temporal grounding.

For each input instance, a video-language instruction pair, the model samples a group of $G$ candidate responses $\{o_1, o_2, \dots, o_G\}$. Each response is evaluated using the appropriate task-specific reward function, yielding a set of scalar rewards $\{R_1, R_2, \dots, R_G\}$. These rewards are then normalized within the group to compute the relative advantage for each sample:

\begin{equation}
\small
A_i = \frac{R_i - \text{mean}(\{R_1, \dots, R_G\})}{\text{std}(\{R_1, \dots, R_G\})},
\end{equation}

where \( A_i \) reflects the relative quality of the $i$-th response. The model is updated to increase the likelihood of higher-scoring responses and reduce that of lower-scoring ones, enabling preference-driven fine-tuning guided by human-verifiable objectives.

By integrating our task-specific temporal rewards with GRPO's relative optimization framework, \emph{Temporal-RLT} provides a unified and scalable post-training strategy. It supports a wide range of video-language tasks, promotes structured and interpretable outputs, and eliminates the need for manually engineered reward critics.

\begin{figure*}[!tp]
\centering
\begin{minipage}{.4\textwidth}
\includegraphics[height=120pt]{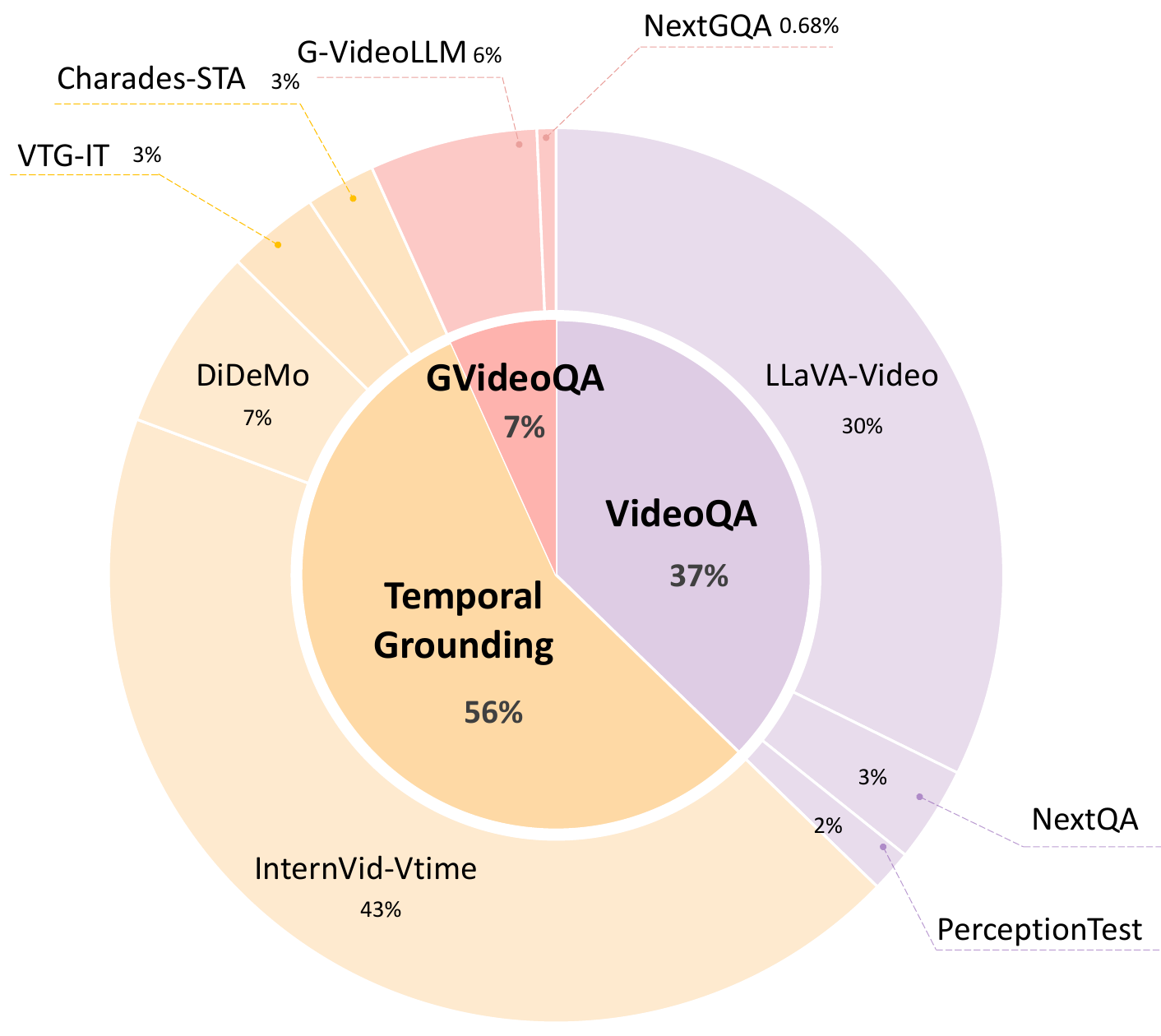}
\end{minipage}%
\hspace{4mm}
\begin{minipage}{.55\textwidth}
\centering
\renewcommand{\arraystretch}{1.18}
\resizebox{0.98\textwidth}{!}{ 
\setlength\tabcolsep{1pt}
\begin{tabular}{c|lc|c|r}
\toprule 
\textbf{Task} & \multicolumn{2}{c|}{\textbf{Data Source}} & \multicolumn{1}{c|}{\textbf{Domain}} & \textbf{Data Size}  \\  
\midrule

\multirow{3}{*}{\textbf{VideoQA}} 
 & & LLaVA-Video~\cite{li2024llava} & Diverse & 158439 \\  
 & & NextQA~\cite{nextqa}& Reasoning & 17024 \\   
 & & PerceptionTest~\cite{patraucean2023perception} & Perception, Reasoning & 7392 \\
\midrule
\multirow{4}{*}{\shortstack{\textbf{Temporal}\\\textbf{Grounding}}} 
 & & InternVid-Vtime~\cite{huang2024vtimellm} & Diverse & 213158 \\   
 & & DiDeMo~\cite{didemo} & Diverse & 33002 \\
 & & VTG-IT~\cite{guo2025vtg}& Diverse & 16318 \\
 & & Charades-STA~\cite{charadesta} & Indoor & 12408 \\  
\midrule
\multirow{2}{*}{\shortstack{\textbf{Grounded}\\\textbf{VideoQA}}} 
 & & G-VideoLLM~\cite{wang2024grounded} & Vlog, News and Activity & 29797 \\
 & & NextGQA~\cite{nextgqa} & Reasoning & 3358 \\
\bottomrule
\end{tabular}
}
\end{minipage}
\vspace{1mm}
\caption{\textbf{Data statistic of Temporal-RLT-Full-490k training dataset.} We find that converting all answers to an open-ended format is critical in reliably assessing question difficulty and effective model training.}
\label{fig:fig3}
\end{figure*}

\section{Rethinking Data Efficiency for Video RLT}
\label{sec:data}

Data efficiency is critical for the post-training and fine-tuning of LLMs and MLLMs. It is equally essential in GRPO-style RLT for video understanding, where effective utilization of supervision is key to further improving model capability. In this section, we focus on the distinct characteristics of GRPO optimization and analyze data efficiency beyond conventional assumptions of data diversity. Building on our dual-reward formulation, we separately analyze the discrete and continuous reward paradigms, and propose a principled strategy to enhance reward-data utilization in the video domain.

\subsection{Difficulty Estimation via Multi-Sampling}

GRPO relies on relative preference signals within sampled groups to guide learning. Therefore, the informativeness of a training sample largely depends on whether its sampled outputs induce sufficient diversity in reward values. If all outputs are either highly correct or uniformly poor, the absence of contrast limits optimization signal. To quantify this, we adopt a unified multi-sampling-based difficulty estimation strategy. For each training instance, we perform $N \gg K$ inference runs using a base VideoLLM (pre-RLT) and compute reward scores across sampled outputs. For discrete tasks (\emph{e.g.}, multi-choice QA), we count the number of correct responses. For continuous tasks (\emph{e.g.}, temporal IoU), we evaluate the internal reward spread. Samples are then categorized based on their observed output variance, forming the basis for subsequent selection and prioritization.

\subsection{Data Efficiency under Discrete Rewards}

In multi-choice QA tasks, each sampled response receives a binary reward $r_i \in \{0, 1\}$ based on whether the predicted answer matches the ground truth. Effective GRPO optimization requires a \emph{mixture} of correct and incorrect answers within the sampled group, so that lower-quality responses can be aligned toward higher-quality ones.

To approximate this mixture potential, we define difficulty based on the frequency of correct predictions among $N$ base model inferences. We label a sample as:
\[
\text{Easy if } c \geq \tau_{\text{easy}}, \quad \text{Hard if } c \leq \tau_{\text{hard}}, \quad \text{otherwise Medium},
\]
where $c$ is the number of correct predictions out of $N$. We retain medium-difficulty samples to ensure sufficient behavioral variance, which supports more effective preference-based learning. 

\subsection{Data Efficiency under Continuous Rewards}

In grounding tasks with continuous temporal IoU supervision, binary correctness no longer applies. Instead, we assess the spread of IoU scores within each sampled group to estimate difficulty. Intuitively, samples whose predictions yield a wide range of IoUs provide stronger relative learning signals. To avoid instability in direct variance computation under limited sampling ($K$), we define an approximate metric based on reward discrepancy:
\[
\Delta_{\text{IoU}} = \max_{i} \text{IoU}_i - \text{mean}_{i}(\text{IoU}_i),
\]
where a higher $\Delta_{\text{IoU}}$ indicates greater internal diversity. Samples with negligible spread are excluded from training, as they fail to generate preference gradients within the group.

\subsection{Difficulty-Aware Data Construction}

To support effective training under GRPO, we construct a difficulty-aware dataset for video RLT, comprising two specialized subsets: (1) Temporal-RLT-Full-490k as our foundational dataset containing diverse domain data for VideoQA, Temporal Video Grounding, and Grounded VideoQA tasks, and (2) Temporal-RLT-32k, a refined subset optimized for sample diversity and training efficacy. Our raw corpus includes a large set of video-question pairs and temporal grounding data with diverse semantics and temporal scopes, covering multiple benchmark datasets as visualized in Fig.~\ref{fig:fig3}.  

We first conduct offline difficulty estimation on Temporal-RLT-Full-490k using the base VideoLLM through large-scale scaling inference. Each training instance is evaluated using our unified multi-sampling pipeline and assigned a task-specific difficulty score (based on accuracy frequency for discrete tasks and intra-group reward variation for continuous tasks). Based on this analysis, we derive Temporal-RLT-32k, a carefully curated training subset composed primarily of medium-difficulty samples that prove empirically more effective in generating stable and informative GRPO updates.

This difficulty-aware data construction improves sample efficiency by reducing redundant updates on uninformative (easy) samples and mitigating instability from noisy (hard) samples. It also ensures consistency across heterogeneous reward types, enabling a unified optimization framework for semantic and temporal reasoning tasks.

\section{Experiment}
In this section, we conduct comprehensive experiments across a range of General VideoQA, Temporal Grounding, and Video Reasoning benchmarks to evaluate the effectiveness of our Temporal-RLT framework. We further present an in-depth empirical analysis of data efficiency in GRPO-based RLT, highlighting the impact of our sample selection and dynamic data strategies.
\subsection{Implementation Details}

\textbf{Training and Inference Details.} All experiments are conducted using the Qwen-VL-2.5-7B~\cite{qwen25vl} model, fine-tuned with GRPO-based reinforcement learning. We sample eight candidate responses per input during training, with a temperature of 1.0 and top-$p$ of 0.99. Training is performed with a batch size of 48 on 8 NVIDIA A100 80GB GPUs. For inference, we adopt conservative decoding settings (temperature = 0.01, top-$p$ = 0.001) to ensure output stability, following the official Qwen-VL-2.5 deployment configuration. Video inputs are processed at 2 FPS using Qwen-VL-2.5’s dynamic resolution module, with at least 4 tokens per frame and a maximum of 2048 tokens per video. Spatial resolution is automatically adjusted while preserving aspect ratio to ensure efficient computation. All training and inference procedures are conducted in BF16 precision for numerical stability.

\textbf{Benchmarks.} To evaluate our method across different aspects of video understanding, we conduct experiments on a diverse set of benchmarks. For general video-language understanding, we use MVBench~\cite{li2024mvbench}, TempCompass~\cite{tempcompass}, and VideoMME~\cite{videomme} (excluding subtitles). For video reasoning, we evaluate on MMVU~\cite{mmvu}, which focuses on multi-choice reasoning tasks. For temporal video grounding, we test on Charades-STA~\cite{charadesta} and ActivityNet~\cite{anetgrounding}, while reasoning-driven temporal grounding is assessed on ActivityNet-RTL~\cite{huang2024lita}. Finally, for grounded video question answering, we report results on the NextGQA test set~\cite{nextgqa}. These benchmarks collectively assess the model’s capabilities in semantic comprehension, temporal localization, and reasoning across diverse tasks.
 
\subsection{Main Results}
As shown in Tab.~\ref{tab:main_results}, our Temporal-RLT approach achieves significant improvements in multiple video understanding tasks compared to standard SFT. In temporal video grounding, our method yields substantial absolute gains: $+14.0$ mIoU on Charades, $+14.7$ on ActivityNet, and $+9.5$ on the reasoning-intensive ActivityNet-RTL benchmark. These results confirm that our RL-based strategy is highly effective at improving temporal localization. Importantly, while SFT often leads to performance degradation on general VideoQA benchmarks due to overfitting or misalignment between pretraining and downstream objectives, our method maintains strong generalization and consistently outperforms the SFT baseline across both general and reasoning-focused tasks. Specifically, we observe a $+3.7$ improvement on the MMVU benchmark, highlighting the model's enhanced reasoning capability under our training paradigm. Moreover, when compared to Video-R1~\cite{videor1}, our method achieves superior performance across all evaluated tasks while requiring less data for training. This demonstrates the effectiveness of our data selection strategy and highlights the data efficiency of our RLT pipeline. Together, these results showcase the superiority of our Temporal-RLT framework in achieving strong performance across grounding, general QA, and reasoning tasks, while maintaining high data efficiency and strong generalization across diverse video benchmarks.

\begin{table*}[t!]
\caption{\textbf{Main Experiment Results.}  Comparison of our Temporal-RLT framework against baseline VideoLMMs and post-training baselines across multiple video understanding and grounding benchmarks.}

\setlength{\tabcolsep}{1.1pt}

\footnotesize
\begin{center}

\resizebox{\textwidth}{!}{%
\begin{tabular}{l| c c c | c c c | c | c c}
\hline
\multirow{3}{*}{\textbf{Method}} & \multicolumn{3}{c}{\textbf{Temporal Video Grounding}} & \multicolumn{3}{c}{\textbf{General VideoQA}} & \textbf{Reasoning QA} & \multicolumn{2}{c}{\textbf{Grounded QA}} \\
\cmidrule(lr){2-4} \cmidrule(lr){5-7} \cmidrule(lr){8-8} \cmidrule(lr){9-10}
& Charades & ANet & ANet-RTL & MVBench & TempCompass & VideoMME & MMVU & \multicolumn{2}{c}{NextGQA} \\
& mIoU & mIoU & mIoU & Avg & Avg & Avg (wo sub) & Avg & mIoU & acc \\ 

\hline
\multicolumn{10}{c}{General VideoLLM} \\ \hline
    LLaMA-VID\cite{li2024llama} & - & - & - & 41.9 & 45.6 & - & - & - & - \\
    VideoLLaMA2\cite{cheng2024videollama} & - & - & - & 54.6 & - & 47.9 & 44.8 & - & - \\
    LongVA-7B\cite{zhang2024long} & - & - & - & - & 56.9 & 52.6 & - & - & - \\
    Video-UTR-7B\cite{yu2025unhackable} & - & - & - & 58.8 & 59.7 & 52.6 & - & - & - \\
    LLaVA-OV-7B\cite{li2024llava} & - & - & - & 56.7 & - & 58.2 & 49.2 & - & - \\
    Kangeroo-7B\cite{liu2024kangaroo} & - & - & - & 61.1 & 62.5 & 56.0 & - & - & - \\ \hline
\multicolumn{10}{c}{\cellcolor{red!5} GRPO-based Method and Baseline
} \\ \hline
\rowcolor{red!5} Qwen-VL-2.5\cite{qwen25vl} &28.0 & 24.0 & 6.0 & 65.3 & 70.9 & 56.1 & 61.3 & 20.2 & 77.2 \\
        
\rowcolor{red!5} Qwen-VL-2.5-SFT & 43.0 & 24.3 & 18.1 & 62.0 & 68.7 & 49.6 & 52.5 & 28.3 & 70.6 \\

\rowcolor{red!5}  Video-R1\cite{videor1} & - & - & - & 62.7 & 72.6 & 57.4 & 64.2 & - & - \\
\rowcolor{red!15} \textbf{\work \ (ours)} & \textbf{57.0} & \textbf{39.0} & \textbf{27.6} & \textbf{68.1} & \textbf{73.3} & \textbf{57.6} & \textbf{65.0} & \textbf{37.3} & \textbf{78.7} \\ \hline
\end{tabular}%
}
\end{center}
\label{tab:main_results}
\end{table*}

\subsection{Ablation Study}
We present a series of ablation studies to understand how task composition, reward design, and data selection affect the performance of our Temporal-RLT framework. The results reveal critical insights into optimization behavior and generalization across video-language tasks under GRPO-based tuning.

\textbf{Multi-Task Ablation.}
Our experiments systematically evaluate the impact of different training data configurations on model performance. The TVG-only setup yields substantial improvements on temporal grounding tasks, with gains of $+27.6$ mIoU on Charades and $+11.6$ on ActivityNet, while maintaining reasoning performance (no degradation on MMVU) and improving grounded QA results ($+5.8$ mIoU). These findings suggest that enhanced temporal understanding can transfer effectively to related tasks. Adding VideoQA data introduces expected trade-offs: while it significantly improves question-answering accuracy on both general and reasoning-oriented VideoQA benchmarks, it leads to a moderate decline in grounding precision. Further augmenting with grounded VQA (G-VQA) produces more nuanced effects, including domain-specific gains on ActivityNet ($+7.6$) and smaller but meaningful improvements on Charades-STA ($+1.1$), along with a reasoning enhancement of $+0.8$ on MMVU. Together, these results highlight the partially decoupled development of temporal understanding and reasoning abilities. They also reveal critical trade-offs in data composition during RLT, illustrating the optimization boundaries across different video understanding tasks.
\begin{table*}[t]
\vspace{1mm}
\tablestyle{5pt}{1.1}
    \begin{center}

    \caption{\textbf{Ablation Study on Multi-Task RL Tuning Components}. We analyze and compare the effects of using different training data and reward types for Temporal-RLT. }
    \resizebox{\textwidth}{!}
{
    \begin{tabular}{@{}l c c c c c c c c c c@{}}
        \toprule
        \multirow{3}{*}{\textbf{Method}} & \multicolumn{3}{c}{\textbf{Temporal Video Grounding}} & \multicolumn{3}{c}{\textbf{General VideoQA}} & \multicolumn{1}{c}{\textbf{Reasoning QA}} & \multicolumn{2}{c}{\textbf{Grounded QA}} \\
        \cmidrule(lr){2-4} \cmidrule(lr){5-7} \cmidrule(lr){8-8}  \cmidrule(lr){9-10}
        & Charades & ANet & ANet-RTL & MVBench & TempCompass & VideoMME  & MMVU &  \multicolumn{2}{c}{NextGQA} \\
        & mIoU & mIoU & mIoU & Avg & Avg & Avg (wo sub) & Avg & mIoU & acc \\
     
        \midrule
        
        Qwen-VL-2.5 (baseline) & 28.0 & 24.0 & 6.0 & 65.3 & 70.9 & 56.1 & 61.3 & 20.2  & 77.2 \\
        
        \midrule
        \quad +w/ TVG & 55.6 & 35.6 & 23.6 & 61.8 & 71.2 & 52.4 & 61.3 & 29.7 & 75.3 \\
        \quad +
        w/ VQA & - & - & -  & 68.1 & 72.5 & \textbf{58.6} & 63.1  & 25.0 & 78.,4 \\
        \quad +w/ VQA  + TVG & 55.9 & 31.3 & 23.2 & 68.0 & 72.9 & 58.4 & 64.2& 27.0 & 78.2 \\
        \quad +w/ VQA  + TVG + G-VQA & \textbf{57.0} & \textbf{39.0} & \textbf{27.6} & \textbf{68.1} & \textbf{73.3} & {57.6} & \textbf{65.0} & \textbf{37.3} & \textbf{78.7}  \\
        
        \bottomrule              
    \end{tabular}
}
 \end{center}
    \label{tab:ablation_studyperformance}
\end{table*}

\begin{table*}[ht]
\vspace{1mm}
\tablestyle{5pt}{1.1}
    \begin{center}

    \caption{\textbf{Ablation Study on More Video Tokens}. We analyze and compare the effects of using more Video Tokens for Inference Result . }
    \resizebox{\textwidth}{!}
{
    \begin{tabular}{@{}l c c c c c c c c c c c@{}}
        \toprule
        \multirow{3}{*}{\textbf{Method}} & \multirow{3}{*}{\textbf{Video Tokens}} & \multicolumn{3}{c}{\textbf{Temporal Video Grounding}} & \multicolumn{3}{c}{\textbf{General VideoQA}} & \multicolumn{1}{c}{\textbf{Reasoning QA}} & \multicolumn{2}{c}{\textbf{Grounded QA}} \\
        \cmidrule(lr){3-5} \cmidrule(lr){6-8} \cmidrule(lr){9-9}  \cmidrule(lr){10-11}
        & & Charades & ANet & ANet-RTL & MVBench & TempCompass & VideoMME  & MMVU &  \multicolumn{2}{c}{NextGQA} \\
        & & mIoU & mIoU & mIoU & Avg & Avg & Avg (wo sub) & Avg & mIoU & acc \\
     
        \midrule
        
        Qwen-VL-2.5 & 2048 & 28.0 & 24.0 & 6.0 & 65.3 & 70.9 & 56.1 & 61.3 & 20.2  & 77.2 \\
        Qwen-VL-2.5 & 4096 & 35.7 & 33.4 & 7.1 & 66.5 & 71.0 & 59.4 & 61.1 & 26.3 & 76.5 \\
        Qwen-VL-2.5-SFT & 2048 & 43.0 & 24.3 & 18.1 & 62.0 & 68.7 & 49.6 & 52.5 & 28.3 & 70.6  \\
        Qwen-VL-2.5-SFT & 4096 & 43.8 & 23.9 & 18.7 & 63.0 & 68.9 & 52.1 & 55.2 & 27.4 & 71.3 \\
        Video-R1 & 2048 & - & - & - & 62.7 & 72.6 & 57.4 & 64.2 & - & - \\
        Video-R1 & 4096 & - & - & - & 63.9 & 73.2 & 59.3 & 63.8 & -  & - \\
        \work \  (ours) & 2048 & {57.0} & {39.0} & {27.6} & {68.1} & {73.3} & {57.6} & {65.0} & {37.3} & {78.7} \\
        \work \  (ours) & 4096 & \textbf{58.0} & \textbf{41.3} & \textbf{33.0} & \textbf{69.0} & \textbf{73.6} & \textbf{59.6} & \textbf{66.1} & \textbf{38.2} & \textbf{79.7} \\

        \bottomrule
    \end{tabular}
}
    \label{tab:abl-videotoken}

 \end{center}
\end{table*}

\textbf{Performance with More Video Tokens.}
As shown in \ref{tab:abl-videotoken}, by increasing the number of input video tokens, we achieve consistent performance improvements across four types of tasks. Moreover, compared to the SFT model, our method benefits more significantly from the increase in video token numbers. In contrast to the previous Video-R1, our approach demonstrates stable performance improvements across all tasks. Notably, when increasing the number of input video tokens, Video-R1 experiences a performance drop on the reasoning QA task in the MMVU benchmark, whereas our method maintains robust gains.

\begin{table*}[t]
\tablestyle{5pt}{1.1}
    \centering
    \caption{\textbf{Ablation Studies: Video QA and TVG Data Selection}. }
    \begin{subtable}{0.54\textwidth}  
        \centering
        \resizebox{\textwidth}{!}{
            \begin{tabular}{@{}c c c c c c c c c c c@{}}
                \toprule
                \multirow{2}{*}{\textbf{Easy: Middle: Hard}} & \multicolumn{3}{c}{\textbf{General VideoQA}} & \multicolumn{1}{c}{\textbf{Reasoning QA}}  \\
                \cmidrule(lr){2-4} \cmidrule(lr){5-6} & \textbf{MVBench} & \textbf{TempCompass} & \textbf{VideoMME}  & \textbf{MMVU}  \\
                \midrule
                4 : 4 : 2 & 64.3 & 70.0 & 52.8 & 59.5   \\
                2 : 4 : 4 & 65.9 & 70.3 & 55.9 & 63.0   \\
                2 : 6 : 2 & 67.2 & 71.3 & 56.8 & 62.1   \\
                1 : 8 : 1 & 68.1 & \textbf{73.4} & 57.1 & \textbf{63.4}  \\
                0 : 10 : 0 & \textbf{68.1} & 72.5 & \textbf{58.6} & 63.1   \\
                \bottomrule
            \end{tabular}
        }
        \vspace{0.5em}  

        \caption{Ablation for Video QA Data Selection}
        \vspace{3mm}
        \label{tab:abl-mc}
    \end{subtable}
    \hspace{0.02\textwidth}  
    \begin{subtable}{0.4\textwidth}
        \centering
        \resizebox{\textwidth}{!}{
            \begin{tabular}{@{}c c c c c @{}}
                \toprule
                \multirow{2}{*}{\textbf{$\Delta_{\text{IoU}}$}} & \multicolumn{4}{c}{\textbf{Charades-STA}} \\
                \cmidrule{2-5} & Recall@0.3 & Recall@0.5 & Recall@0.7 & mIoU \\
                \midrule
                0 & 78.2 & 63.9 & 37.4 & 54.7    \\
                0.1 & 78.0 & 64.8 & 38.9 & 54.9  \\
                0.2 & \textbf{78.8} & 63.7 & 38.5 & 55.0   \\
                0.3 & 78.6 & \textbf{64.5} & \textbf{39.9} & \textbf{55.5}   \\
                \bottomrule
            \end{tabular}
        }
        \vspace{0.5em}  
        \caption{Ablation for TVG Data Selection}
        \vspace{3mm}
        \label{tab:abl-tvg}
    \end{subtable}
    
\end{table*}
\textbf{Data Selection for Multi-choice VideoQA.} The ablation study investigates the impact of difficulty distribution in multi-choice training data. As shown in Tab.~\ref{tab:abl-mc}, reducing the proportion of easy samples while increasing medium-difficulty data improves both general and reasoning performance (evident from the comparison between the first and third rows). Similarly, increasing hard samples while reducing easy ones, while keeping medium data fixed, also enhances performance (first vs. second row), suggesting that easy data contribute little to effective training. Rows two through five further show that gradually increasing the share of medium-difficulty samples, while decreasing both easy and hard samples, leads to consistent improvements across VideoQA benchmarks. Notably, comparing the second and third rows reveals that while a higher proportion of hard data can benefit reasoning, it may negatively impact general performance. These results highlight the importance of balancing difficulty levels, particularly by increasing medium-difficulty samples, to optimize model training under GRPO-based RLT.

\textbf{Data Selection for Temporal Grounding.} In temporal video grounding, the reward signal is continuous, making intra-group variability a critical factor for effective learning. We use $\Delta_{\text{IoU}}$ as a proxy for internal diversity within sampled outputs. As shown in Tab.~\ref{tab:abl-tvg}, increasing $\Delta_{\text{IoU}}$ from $0$ to $0.3$ consistently improves grounding performance. These results indicate that samples with greater prediction variance provide more informative preference signals, making them more suitable for GRPO-based RLT in temporal grounding tasks.

\begin{table*}[t]
\tablestyle{5pt}{1.1}
    \centering
    \caption{\textbf{Temporal Video Grounding OOD Evaluation}.}
    \resizebox{\textwidth}{!}
{
    \begin{tabular}{@{}c c c c c c c c c c c c c c c @{}}
        \toprule
       \multirow{2}{*}{\textbf{Tuning Type}} & \multicolumn{4}{c}{\textbf{Charades-STA}} & \multicolumn{4}{c}{\textbf{ActivityNet}} & \multicolumn{4}{c}{\textbf{ActivityNet-RTL}} \\
        \cmidrule(lr){2-5} \cmidrule(lr){6-9} \cmidrule(lr){10-13} &  R@0.3 & R@0.5 & R@0.7 & mIoU & R@0.3 & R@0.5 & R@0.7 & mIoU & R@0.3 & R@0.5 & R@0.7 & mIoU \\
        \midrule
        \ding{55} & 42.4 & 29.8 & 14.0 & 28.0 & 34.4 & 22.5 & 11.6 & 24.0 & 7.9 & 2.6 & 2.9 & 6.0    \\
        SFT & 73.9 & 61.6 & 38.5 & 52.8 & 33.4 & 18.9 & 9.0 & 23.1 & 24.0 & 14.8 & 7.4 & 17.8  \\
        RLT & 80.2 & 68.3 & 44.5 & 57.9 & 56.9 & 38.4 & 20.2 & 39.1 & 40.2 & 22.7 & 10.9 & 26.3  \\
        
        \bottomrule
    \end{tabular}
}
    \label{tab:ood}
\end{table*}

\textbf{Generalization of RLT Across Video Tasks.} To assess the generalization capability of RLT in video understanding, as shown in \ref{tab:ood}, we perform both SFT and GRPO-based RLT using only the Charades training set, and evaluate performance on both in-domain and out-of-domain benchmarks. Compared to the SFT baseline, the GRPO-trained model demonstrates substantial performance gains and stronger reasoning capabilities, as evidenced by the generation of explicit "think" traces. Specifically, on the in-domain Charades dataset, our model achieves a $+4.7$ mIoU improvement. More notably, it generalizes effectively to out-of-domain settings, with a $+16.0$ mIoU gain on ActivityNet and a $+8.3$ mIoU gain on the reasoning-focused ActivityNet-RTL benchmark. These results highlight not only the effectiveness of our approach for temporal localization, but also its ability to transfer reasoning capabilities across tasks, thereby enhancing both localization accuracy and temporal reasoning performance. This demonstrates that GRPO-based training equips the model with stronger generalization and adaptability to diverse video domains and task requirements.

\begin{figure}[h]
   \centering
   \includegraphics[width=1\textwidth]{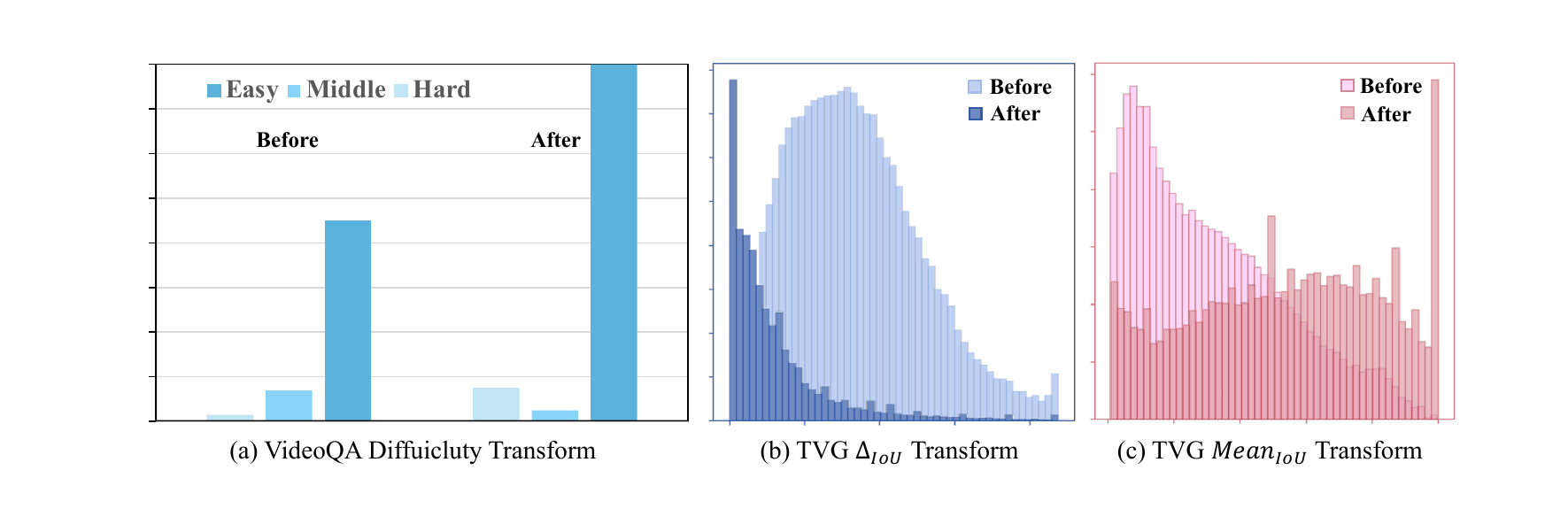} 
   \caption{\textbf{Changes in data distribution before and after RLT.} We report the distribution of correctness counts for MC-VideoQA and the IoU variance (max–min) for temporal grounding. The post-RLT data reflects only the subset of samples selected from the original dataset based on our sampling strategy. }
   \label{fig:distribution}
\end{figure}

\textbf{Distributional Shifts Before and After RLT.} We further examine changes in the model's output distribution induced by RLT training as shown in \ref{fig:distribution}. For the VideoQA task, we observe that prior to RLT, most samples were classified as easy, though a substantial portion exhibited medium or hard difficulty. After RLT, the medium-difficulty samples largely shifted toward either easy or hard, indicating a loss in output diversity. For the temporal grounding task, we compare changes in the distribution of the $\Delta_{\text{IoU}}$ (max–min) and mean IoU scores. Before training, $\Delta_{\text{IoU}}$ peaked near the center, while mean IoU was skewed leftward. Post-RLT, the $\Delta_{\text{IoU}}$ peak shifts closer to zero, and the mean IoU distribution moves rightward. This suggests an overall performance improvement, but also a reduction in intra-group variance. These trends reveal important task-specific implications. In the case of VideoQA, the convergence of predictions implies that further RLT provides diminishing returns, as the model's output becomes overly deterministic and less capable of generating meaningful preference gradients. In contrast, temporal grounding tasks retain sufficient prediction variance post-RLT, indicating continued room for improvement. This highlights the importance of considering output diversity in evaluating the saturation and potential of RLT for different video tasks.

\vspace{-1mm}
\section{Related Work}
\vspace{-1mm}

\textbf{Video Large Language Models.} Video Large Language Models (VideoLLMs)~\cite{qwen25vl, li2024llava, ye2024mplug, zhang2024long, li2025llava} have emerged as unified architectures for video-language understanding. Early works~\cite{li2023videochat, li2024llama} focused on tasks such as video captioning, video question answering, and video-grounded dialogue, typically trained via supervised objectives. More recent efforts~\cite{huang2024vtimellm, liu2024st, wang2024hawkeye,huang2025unleashing,zeng2024timesuite, wang2024grounded, timechat, li2025llava, videoespresso} have introduced temporally structured datasets and architectures to better capture frame-level alignment and long-range dependencies. Techniques such as sparse sampling and temporal token compression have been proposed to improve temporal perception and efficiency. In parallel, temporal reasoning has gained increasing attention. Lita~\cite{huang2024lita} introduced a benchmark targeting temporal grounding, while Momentor~\cite{qian2024momentor} released a large-scale dataset for temporal reasoning. However, most approaches remain grounded in supervised fine-tuning (SFT), with reinforcement learning tuning (RLT) still underexplored for developing temporal understanding and generalizable reasoning in VideoLLMs.

\textbf{Reinforcement Learning Tuning for MLLMs.} Reinforcement learning (RL) has proven effective for improving alignment and reasoning in large language and multimodal models. DeepSeek-R1~\cite{guo2025deepseek, shao2024deepseekmath} showed that preference-based tuning with outcome-level rewards can enable strong chain-of-thought (CoT) reasoning without step-level supervision. Building on this, Kimi-k1.5~\cite{team2025kimi15} adopted rule-based RL strategies to improve reasoning across both text and vision modalities. In the multimodal setting, RLT with verifiable rewards~\cite{visualrft, r1vl, visonr1, zhou2025r1} has been used to enhance visual reasoning in tasks such as VQA and image grounding. However, applying RL to video understanding remains limited due to challenges in designing temporally sensitive rewards and modeling spatiotemporal consistency. Only a few works, such as Video-R1~\cite{videor1}, have explored this direction, leaving significant space for further research on reward formulation and training efficiency in the video domain.

\vspace{-1mm}
\section{Conclusion}
\vspace{-1mm}
In this paper, we present an RL-based post-training framework for enhancing the video-specific reasoning abilities of LLMs. We build upon the GRPO algorithm and propose a dual-reward formulation that explicitly supervises both semantic and temporal reasoning. To ensure effective optimization under this framework, we introduce a variance-aware sample selection method based on repeated inference and a dynamic sampling strategy that progressively refines the training set. Through the construction of two dedicated datasets and comprehensive experiments on a wide range of video understanding tasks, we demonstrate that our approach significantly improves model performance in reasoning-centric scenarios, even when compared to existing SFT and RLT baselines. Our results highlight the importance of tailored reward design and data selection in post-training VideoLLMs and open up new directions for scalable, reasoning-driven video understanding.

\textbf{Limitations.} This work focuses on reward formulation and data efficiency within the GRPO-based RLT for video understanding. We do not explore the construction of high-quality, fine-grained thinking traces to support SFT for enhancing the model’s explicit thinking and reasoning abilities. Incorporating such structured thinking supervision remains an important direction for future research.


{
\bibliographystyle{plain}
\bibliography{reference}
}

\newpage
\appendix

\section{Data Selection Implementation Details}

To filter high-quality data suitable for GRPO training from the Temporal-RLT-Full-490k dataset, we first categorize the data into three types: tasks requiring \textbf{Discrete Rewards}, tasks requiring \textbf{Continuous Rewards}, and tasks requiring both. These categories align with grounded VideoQA tasks. Initially, we utilize the entire dataset to perform \textbf{"No-Think" inference} using the Qwen-VL-2.5-7B model to generate preliminary results. Subsequently, we apply task-specific processing for each category, using the same settings as the GRPO inference process (temperature = 1.0, top-p = 0.99, and sample-N = 8).

\paragraph{Discrete Rewards Data Selection} For discrete rewards data, we set thresholds of \(\tau_{\text{easy}} = 1\) and \(\tau_{\text{hard}} = 7\) to classify the data into easy, medium, and hard categories. Based on our ablation study, which showed that medium-level data performs best for multi-choice tasks, we select 16k samples from the medium category. Additionally, we ensure an even distribution of samples from each data source within the dataset.

\paragraph{Continuous Rewards Data Selection}
For continuous rewards data, our ablation study revealed that a larger \(\Delta_{\text{IoU}}\) leads to better results. Therefore, we select data with \(\Delta_{\text{IoU}} \geq 0.3\). From this subset, we extract 8k samples while maintaining an even distribution of samples across all data sources.

\paragraph{Grounded QA Data Selection}
For tasks requiring both types of rewards, we apply the aforementioned rules simultaneously. However, due to the limited availability of data in this category, we lower the \(\Delta_{\text{IoU}}\) threshold to 0.1. From this filtered subset, we select 8k samples, ensuring an even distribution across all data sources.

By following these steps, we ensure the selection of high-quality, balanced data tailored for GRPO training, optimizing performance across various task types. As a result, we constructed the \textbf{Temporal-RLT-32k} dataset, which serves as a refined subset designed to enhance the training process and improve task-specific outcomes.

\section{Results}

\subsection{Treating VideoQA as Grounded QA}
\begin{table}[t]
\tablestyle{5pt}{1.1}
\centering
\caption{\textbf{Result for Treating VideoQA as Grounded QA}.}
\resizebox{0.6\linewidth}{!}{ 
    \resizebox{\linewidth}{!}{ 
        \begin{tabular}{@{}c c c c c@{}}
            \toprule
            \multirow{2}{*}{\textbf{Infer as GQA}} & \multicolumn{3}{c}{\textbf{General VideoQA}} & \textbf{Reasoning QA} \\
            \cmidrule(lr){2-4} \cmidrule(lr){5-5}
            & {MVBench} & {TempCompass} & {VideoMME} & {MMVU} \\
            \midrule
            \ding{55} & 68.1 & 73.3 & 57.6 & 65.0 \\
            \ding{51} & 67.8 & 72.8 & 57.3 & 63.5 \\
            \bottomrule
        \end{tabular}
    }
}
\label{tab:abl-vqa-as-gqa}
\end{table}

Our model inherently supports treating VideoQA as a form of Grounded QA reasoning, which can be achieved simply by adjusting the prompt format. We evaluated the approach of treating VideoQA as Grounded QA, and while there is some performance degradation as shown in \ref{tab:abl-vqa-as-gqa}, this method allows the model to specify highly relevant time intervals being observed before answering the question.

\subsection{Visualization}
\begin{figure}[t]
   \centering
   \includegraphics[width=1.0\textwidth]{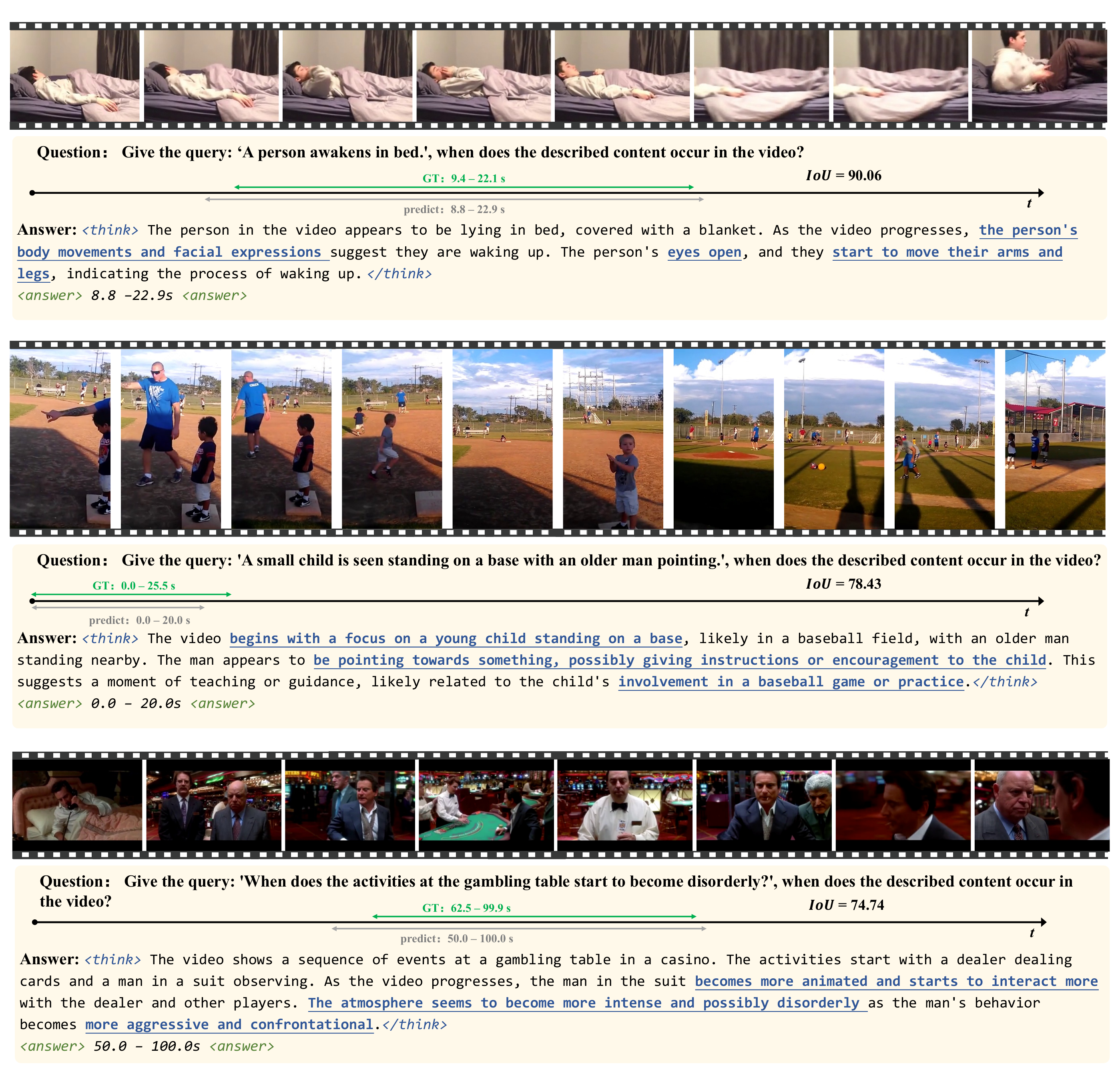} 
   \caption{\textbf{Visualization of Temporal-RLT's reasoning outputs on the Temporal Video Grounding task.} The three examples, from top to bottom, come from Charades-STA, ActivityNet, and ActivityNet-RTL, respectively.}

   \label{fig:vis-tvg}
\end{figure}

\begin{figure}[t]
   \centering
   \includegraphics[width=1.0\textwidth]{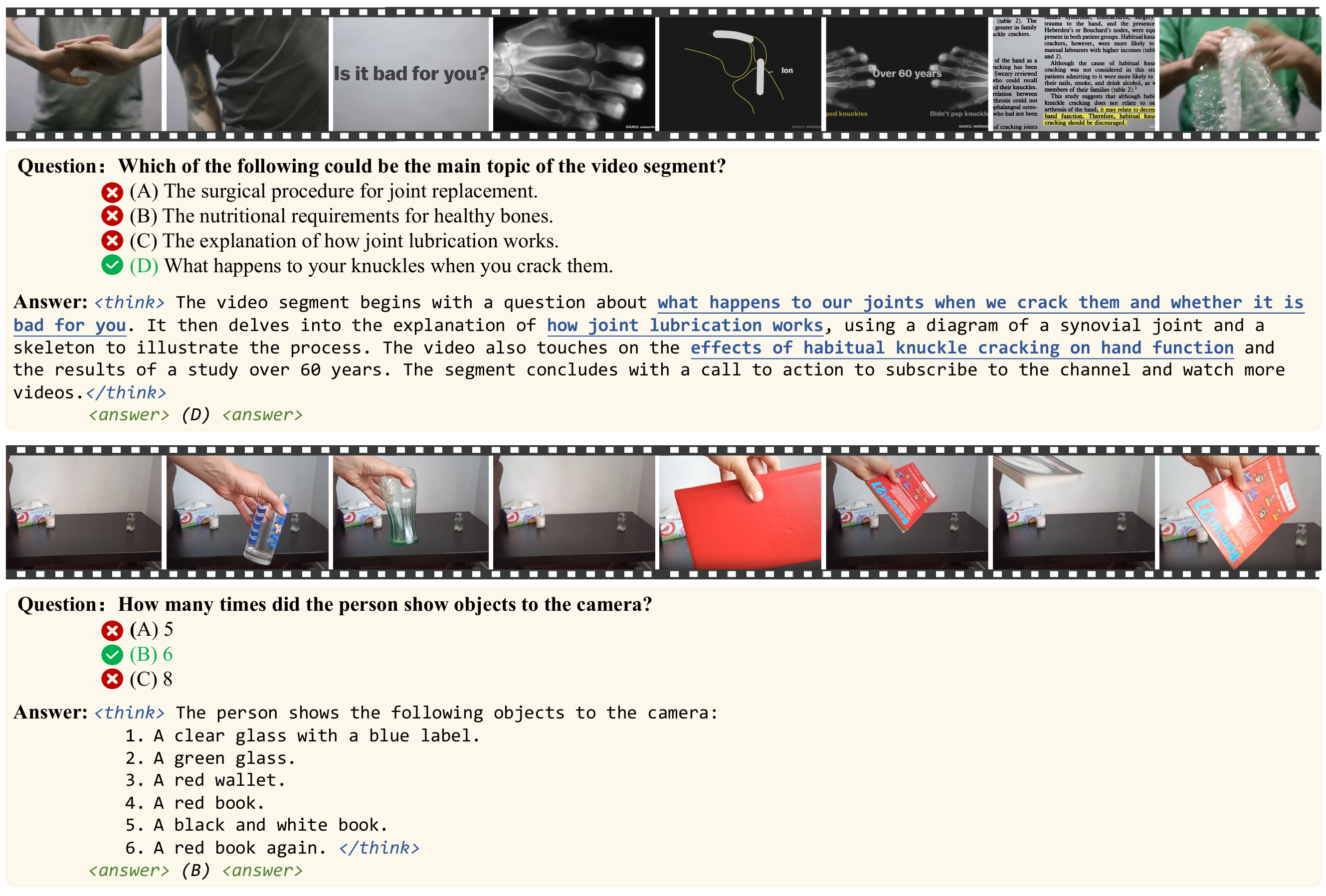} 
   \caption{\textbf{Visualization examples of Temporal-RLT’s reasoning output on Video QA task.} The two examples are from VideoMME and MVBench, respectively.}

   \label{fig:vis-mc}
\end{figure}

\begin{figure}[t]
   \centering
   \includegraphics[width=1.0\textwidth]{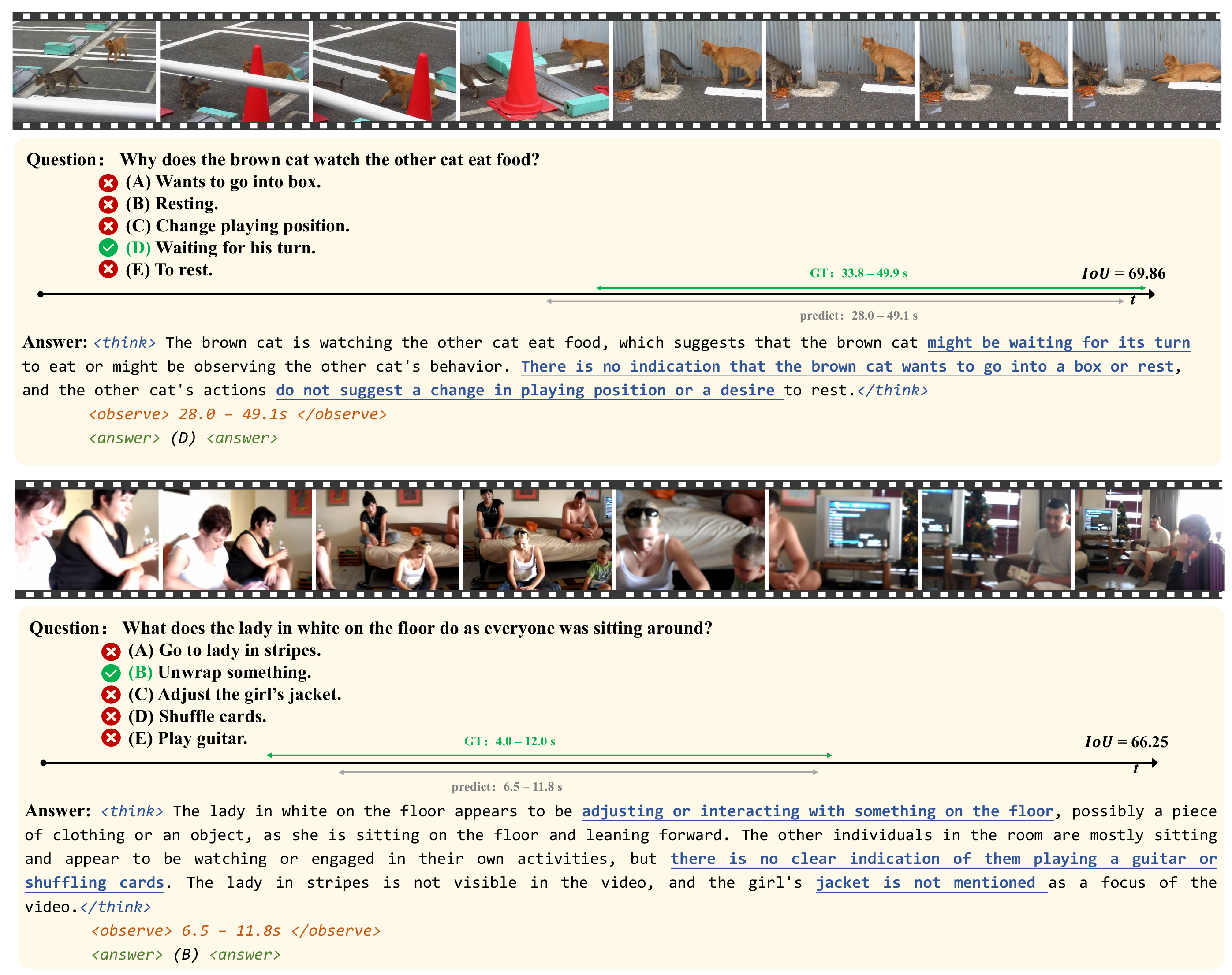} 
   \caption{\textbf{Visualization of Temporal-RLT’s reasoning output on Grounded QA task.} The examples are from the test set of NextGQA.}

   \label{fig:vis-gqa}
\end{figure}

We visualized the example output results on the tasks of Temporal Video Grounding  \ref{fig:vis-tvg}, Video QA \ref{fig:vis-mc}, and Grounded QA \ref{fig:vis-gqa}, including both the reasoning process and the final answers.

\end{document}